\documentclass[conference]{IEEEtran}
\IEEEoverridecommandlockouts
\usepackage{cite}
\usepackage{amsmath,amssymb,amsfonts}
\usepackage{algorithmic}
\usepackage{graphicx}
\usepackage{textcomp}
\usepackage{xcolor}
\usepackage{hyperref}
\usepackage{algorithm,algorithmic}

\def\BibTeX{{\rm B\kern-.05em{\sc i\kern-.025em b}\kern-.08em
		T\kern-.1667em\lower.7ex\hbox{E}\kern-.125emX}}
\begin{document}
	
\title{Feature-Aware Noise Contrastive Learning for Unsupervised Red Panda Re-Identification}
    
\author{
    \author{
	\IEEEauthorblockN{
		Jincheng Zhang,
		Qijun Zhao\IEEEauthorrefmark{1},
		Tie Liu} 
	\IEEEauthorblockA{College of Computer Science, Sichuan University, Chengdu, China\\Email: zhang233@stu.scu.edu.cn, qjzhao@scu.edu.cn, 2022326045009@stu.scu.edu.cn}
} 

}   

	\maketitle
	\begin{abstract}
	To facilitate the re-identification (re-ID) of individual animals, existing methods primarily focus on maximizing feature similarity within the same individual and enhancing distinctiveness between different individuals. However, most of them still rely on supervised learning and require substantial labeled data, which is challenging to obtain. To avoid this issue, we propose Feature-Aware Noise Contrastive Learning (FANCL) method to explore an unsupervised learning solution, which is then validated on the task of red panda re-ID. FANCL designs  a Feature-Aware Noise Addition module to produce noised images that conceal critical features, and employs two contrastive learning modules to calculate the losses. Firstly, a feature consistency module is designed to bridge the gap between the original and noised features. Secondly, the neural networks are trained through a cluster contrastive learning module. Through these more challenging learning tasks, FANCL can adaptively extract deeper representations of red pandas. The experimental results on a set of red panda images collected in both indoor and outdoor environments prove that FANCL outperforms several related state-of-the-art unsupervised methods, achieving high performance comparable to supervised learning methods.
	\end{abstract}
	
	
	\section{Introduction}
	\let\thefootnote\relax\footnotetext{$^{*}$ Qijun Zhao is corresponding author.}
        Deep learning has been applied in the field of animal protection and management, demonstrating its superiority in various aspects such as animal detection and counting~\cite{shao2020cattle}, species recognition~\cite{huang2020interpretable}, individual identification~\cite{wang2021giant}, and behavior analysis~\cite{mathis2020deep}. This paper focuses on the task of animal re-identification (re-ID), aiming to match different images of the same animal individual. Some studies~\cite{huang2020interpretable,wang2021giant,li2019atrw,he2019distinguishing,wang2022two,li2023automatic,zhang2021yakreid,liu2019pose} have utilized deep learning methods to extract discriminative features of individual animals. For instance, Li et al.~\cite{li2019atrw} used keypoints to model different body parts of tigers, while Wang et al.~\cite{wang2021giant} proposed a local patch detector for feature recognition in giant pandas. Although these methods have achieved positive results, they largely depend on supervised learning. Obtaining accurate identity labels for rare animal images in practice is a challenge. Therefore, it inspires us to explore an unsupervised method for animal re-identification, particularly red panda re-identification.

	\begin{figure}[h]
		\centerline{\includegraphics[width=1\linewidth]{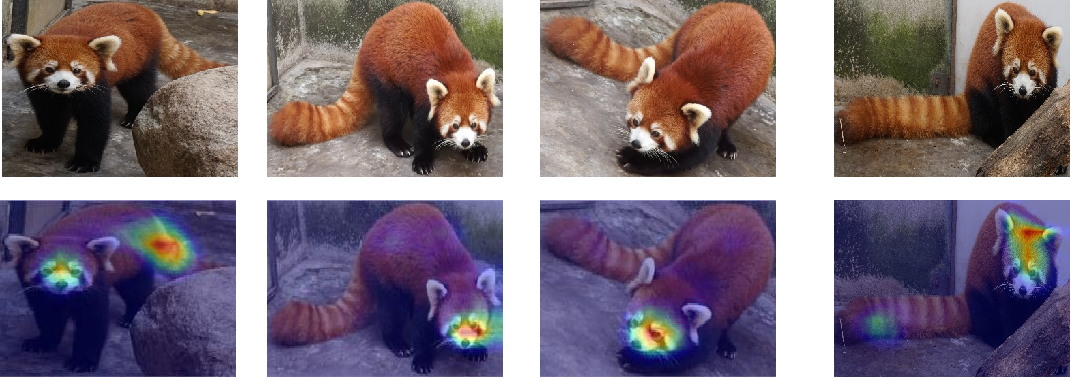}}
		\caption{The Top row shows one red panda in different poses and backgrounds. The bottom row visualizes their corresponding Grad-CAM~\cite{selvaraju2017grad} attention maps obtained by the ResNet50~\cite{he2016deep} baseline model trained in a supervised manner.
		}
		\label{fig1}
	\end{figure}
	
	\begin{figure*}[t]
		\centerline{\includegraphics[width=1\linewidth]{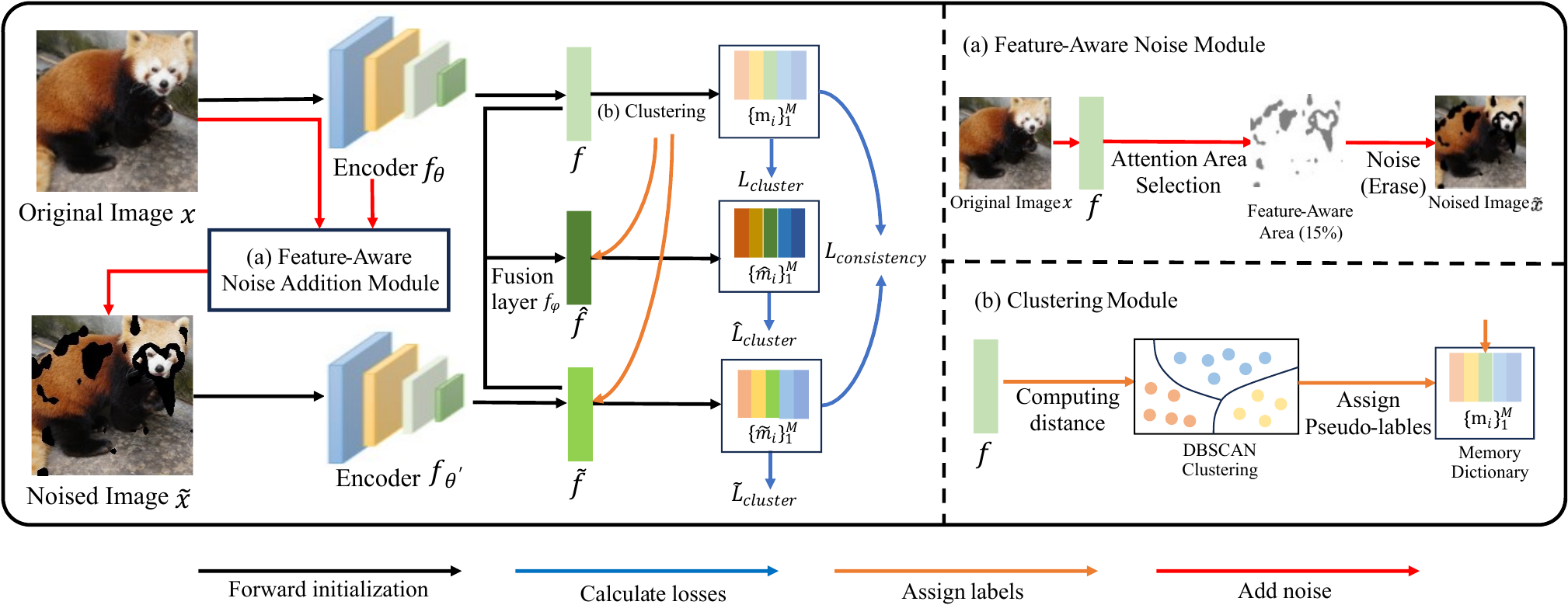}}
		\caption{The framework of our proposed method consists of four modules: Feature-Aware Noise Addition module, Forward Initialization module, Clustering module, and Contrastive Loss Calculation module. The Feature-Aware Noise Addition module (a) obtains a noised image by selecting feature-aware regions from the input image based on the activation map. In the Clustering module (b), pseudo-labels are assigned by clustering the original features on an unlabeled dataset. The Forward Initialization module obtains features for each input and fused features, and finally, the model is trained based on cluster contrastive learning loss and consistency contrastive learning loss. 
		}
		\label{fig2}
	\end{figure*}

	In the field of human and vehicle re-ID, unsupervised learning methods have been extensively studied. These methods can be broadly classified into two categories: one is Unsupervised Domain Adaptation (UDA)~\cite{liu2019adaptive,bak2018domain,wang2018transferable}, which adjusts a model from a labeled source domain to an unlabeled target domain. The other relies on Unsupervised Learning (USL)~\cite{cho2022part,chen2021ice,dai2022cluster} to learn representations purely from unlabeled images. Considering the lower dependency of USL methods on labeled data, we attempt to explore USL to implement red panda re-identification. Specifically, we adopt a widely used two-stage process: firstly, using clustering~\cite{ester1996density,macqueen1967some} to generate pseudo-labels, and then training the re-ID model with these pseudo-labels.
 
	In USL methods, the embedding vectors generated by feature extractors play a key role in measuring image similarities and generating pseudo-labels. Therefore, many methods focus on enabling encoders to generate more robust and representative features. For example, Cluster Contrast~\cite{dai2022cluster} designed a mechanism for updating cluster average features to alleviate the inconsistency issue of features among different images of the same category. PPLR~\cite{cho2022part} proposed a partial feature-based pseudo-label method that focuses on both overall features and local features of horizontally sliced images for finer granularity information. However, for the red panda re-ID task, due to the complex postural variations of red pandas (see Figure~\ref{fig1}), generating effective features is quite challenging.
	
    Inspired by the above ideas, this paper proposes a novel Feature-Aware Noise Contrastive Learning (FANCL) method. As shown in Figure~\ref{fig1}, traditional ResNet-50~\cite{he2016deep} based models often only enable neural networks to focus on certain areas of red panda images (such as the face and tail). Our motivation is to guide feature extractors to learn deeper and more comprehensive feature representation actions. The dual-branch network framework of FANCL is shown in Figure~\ref{fig2}. We feed an original image and a noised image generated by the Feature-Aware Noise Addition (FANA) module. The features of each branch are extracted by independent encoders and then combined in a fusion layer. We construct a multi-cluster memory dictionary for different clusters, where each key corresponds to a cluster's feature vector. Initially, every cluster feature is set as the average feature of all images in that cluster and is updated using a momentum strategy during training. Subsequently, we use a cluster contrastive learning module to calculate the loss between cluster features and query instance features, and a consistency contrastive learning module to narrow the gap between noised features and original features, further prompting the network to learn deeper features of red pandas. The main contributions of this study are as follows:
	\begin{itemize}
		\item To our knowledge, we are the first to apply unsupervised learning methods to animal re-ID, particularly red panda re-ID in this paper. 
		\item An image feature-aware noise addition strategy is proposed, which adds noise (pepper noise in this paper) to key image areas perceived by the neural networks, encouraging the feature extractors to pay broader attention to global feature information.
		\item We elaborately construct an end-to-end contrastive learning framework, which enhances the network's ability to extract robust features by mapping both original and noise-added images to similar feature spaces.
		\item Experiments conducted on a red panda dataset demonstrate that FANCL surpasses several advanced methods in the unsupervised domain and achieves performance comparable to traditional supervised methods.
	\end{itemize}
	
    The rest of this paper is arranged as follows. Section~\ref{related works} reviews related works. Section~\ref{method} presents our method. Section~\ref{experiment} then reports our experimental results. Section~\ref{conclusions} finally concludes the paper with a discussion on future work.
 
    \section{Related Works}
    \label{related works}

    \subsection{Animal re-Identification}
	
	Deep learning methods have been increasingly utilized for individual animal recognition tasks~\cite{huang2020interpretable,wang2021giant,li2019atrw,he2019distinguishing,wang2022two,li2023automatic,zhang2021yakreid,liu2019pose}. Given that individuals of the same species often exhibit only subtle differences in appearance, numerous studies have shifted their focus to leveraging fine-grained local features to enhance recognition performance. For example, He et al.~\cite{he2019distinguishing} developed an automated approach for distinguishing different red pandas by focusing on facial features. Li et al.~\cite{li2019atrw} proposed a precise posture part modeling method for tiger re-identification, while Wang et al.~\cite{wang2022two} designed a lightweight two-stage convolutional neural networks for pig face recognition. Zhang et al.~\cite{zhang2021yakreid} verified the feasibility of Pose-guided Complementary Features Learn and Part-based Convolutional Baseline in yak re-identification. Liu et al.~\cite{liu2019pose} proposed an auxiliary task based on pose classification and designed a multi-branch complementary feature learning neural networks for tiger re-identification. Despite the promising performance of these methods, it is highly demanded to develop unsupervised animal re-ID methods to efficiently explore the unlabeled image data, which is still an open issue in the field of animal recognition.
	
    \subsection{Unsupervised Person re-ID}
	
	Unsupervised person re-identification methods generally fall into two main categories. The first is Unsupervised Domain Adaptation (UDA) methods~\cite{liu2019adaptive,bak2018domain,wang2018transferable}, which transfer knowledge from labeled source data to unlabeled target data. However, these methods still rely on annotations for some data. The second category is pure Unsupervised Learning (USL)~\cite{cho2022part,chen2021ice,dai2022cluster} for person re-identification, independent of any identity labels, making it more suitable for handling animal data where labels are difficult to obtain. In this paper, we focus on the USL approach. Most recent USL-based person re-ID methods concentrate on framework design and pseudo-labels refinement. SpCL~\cite{ge2020self} employs adaptive learning to gradually create more reliable clusters. ICE~\cite{chen2021ice} enhances contrastive learning by utilizing pairwise similarities between instances. Cluster-Contrast~\cite{dai2022cluster} maintains cluster-level features and computes contrastive loss to ensure consistency within each cluster. PPLR~\cite{cho2022part} introduced a pseudo-label method based on partial features, focusing on both the overall and local features of horizontally sliced images for more granular information. Our method in this paper emphasizes the importance of encoders' feature extraction capabilities to make the model's representation of features more comprehensive and discriminative, thereby better dealing with large deformations of animal images.
	
    \section{Proposed Method}
    \label{method}
    As shown in Figure~\ref{fig2}, our FANCL is a dual-branch network: $f_{\theta}$ and $f_{\theta^{\prime}}$, which respectively take an original image $ x $ and a noised image $\widetilde x$ as input. The noised images $\widetilde X = \{\widetilde x_1,\widetilde x_2,\cdots, \widetilde x_N\} $ are obtained by feeding the original images $ X = \{ x_1,\ x_2,\cdots, x_N\} $ to the Feature-Aware Noise Addition (FANA) module, $\widetilde x =\operatorname{FANA}(x) $. We design a feature fusion layer $f_{\varphi}$ to obtain fused features. For simplicity, the output features of $f_{\theta}(x_i)$ and $f_{\theta^{\prime}}(\widetilde x_i)$ are denoted as $f_i$ and $\widetilde f_i$, and the output of the fusion layer $f_{\varphi}(x_i,\widetilde x_i)$ is represented as $\hat{f_i}$. 
    
    Similar to existing cluster-based methods~\cite{cho2022part,dai2022cluster,zhang2021refining}, our approach alternates between clustering and training phases, as shown in Algorithm~\ref{alg1}. During the clustering phase, FANCL generates clusters according to the original features \( f \) and assigns the same pseudo-labels to the features \( \widetilde{f}_i \) and \( \hat{f}_i \) as those of \( f_i \). Then, FANCL maintains three instance memory banks $ {\operatorname{M}} = \{m_1,m_2,\cdots,m_M\} $ ,$\widetilde {\operatorname{M}} = \{\widetilde m_1,\widetilde m_2,\cdots,\widetilde m_M\} $, and $\hat {\operatorname{M}} = \{\hat m_1,\hat m_2,\cdots,\hat m_M\} $. Every memory bank contains $M$ categories, corresponding to cluster features with the same pseudo-labels. In the training stage, FANCL calculates a clustering query loss function inspired by InfoNCE~\cite{oord2018representation}, and a novel feature consistency loss function. Finally, it updates cluster features based on the query features and optimizes each feature extractor through the losses.
    
 	\begin{algorithm}
	\caption{Unsupervised learning pipeline with FANCL}
	\begin{algorithmic} [1]
		\REQUIRE Unlabeled training data \(X\)
		\REQUIRE Initialize the backbone encoder $f_{\theta}$ and $f_{\theta^{\prime}}$ with ImageNet-pretrained ResNet-50
		\REQUIRE Temperature $\tau$ for Eq.~\ref{eq4}
		\REQUIRE Momentum $\alpha$ for Eq.~\ref{eq9}
		\FOR{\( n \) in \( [1, \text{num\_epochs}] \)} 
            \STATE Generate noised images $\widetilde X$ with $\operatorname{FANA}(x) $ in Eq.~\ref{eq3}
		    \STATE Extract feature vectors \( f \), \( \widetilde{f} \) from \( X \), \( \widetilde{X} \) by \( f_{\theta} \), \( f_{\theta^{\prime}} \) and obtain \(\hat f \) through fusion layer $f_{\varphi}$
            \STATE Clustering $f$  with DBSCAN and assign pseudo-labels to \( \widetilde{f} \) and \(\hat f \) 
		\STATE Initialize memory dictionary $ {\operatorname{M}} $, $ \widetilde {\operatorname{M}} $ and $ \hat {\operatorname{M}} $ with Eq.~\ref{eq8}
		\FOR{\( i \) in \( [1, \text{num\_iterations}] \)}
		\STATE Select query instances
        \STATE Extract query feature vectors \( f_q \), \( \widetilde{f_{q}} \) and \(\hat f_q \) 
         \STATE Compute Contrastive Learning loss with Eq.~\ref{eq10}
		\STATE Update cluster features with Eq.~\ref{eq9}
		\STATE Update the encoders by optimizer
		\ENDFOR
		\ENDFOR
	\end{algorithmic}
        \label{alg1}
	\end{algorithm}
	
	\subsection{Feature-Aware Noise Addition Module}
	The Feature-Aware Noise Addition module implements the progress that transitions from an original image to a noised image through three steps. Initially, in the feature activation mapping stage, the original image \(x\) is transformed into a feature map \(A(x)\) through a convolution operation, then passed through an activation function to introduce non-linearity, and resized to match the dimensions of the original image, highlighting significant features within the image. This process is defined as follows:
	\begin{equation}
		A(x) = \text{Resize}\left(\sigma\left(W \ast x + b\right), \text{size}(x)\right)
		\label{eq1}
	\end{equation}
	Following this, in the Noise Position Selection step, a threshold \(\gamma\) binarizes the feature map \(A(x)\), creating a noise position map \(P(x)\) which indicates the areas \(P(x)=1\) where noise should be added. This process is defined as follows:
	\begin{equation}
         P(x) = \begin{Bmatrix}
         1_{\{A(x) \ge \gamma\}}\\
         0_{\{A(x) < \gamma\}}
        \end{Bmatrix}
		\label{eq2}
	\end{equation}
	Finally, in the Noised Image Generation phase, the original image is combined with a noise distribution function \(Noise(x)\), where noise is only added to the areas where \(P(x) = 1\), while the regions where \(P(x) = 0\) retain the original pixel values, resulting in the creation of the ultimate noised image \(\widetilde{x}\). The noise distribution function \(Noise(x)\) used in this paper is pepper noise, which sets the contaminated pixel values to 0. This process is defined as follows:
	\begin{equation}
		\widetilde{x} = x \odot (1 - P(x)) + Noise(x) \odot P(x)
		\label{eq3}
	\end{equation}
	 The entire process ensures that noise can be added to the parts of visual information deemed crucial by the model, thereby enabling the design of FANA to effectively enhance the model's capability to learn deeper-level features.
	
	\subsection{Contrastive Learning Losses}
	\textbf{Cluster Contrastive Learning Loss.} 
	We apply contrastive learning to uncover hidden information within cluster structures. When dealing with original images, for each query instance with feature $f_q$, we treat the cluster feature $m_i$ with the same pseudo-label as a positive sample $m_+$, and the features of all other clusters as negative samples. We define the cluster contrastive learning loss as follows:
	\begin{equation}
		L_{\text {cluster}}=-\log \frac{\exp \left(sim\left(f_q \cdot m_{+}\right) / \tau\right)}{\sum_{m=1}^{M} \exp \left(sim\left(f_q \cdot m \right) / \tau\right)}
		\label{eq4}
	\end{equation}
	$\operatorname{sim}( \mu \cdot \upsilon)$ denotes the cosine similarity between two features $\mu$ and $ \nu$. Its calculation formula is as follows:
	\begin{equation}
		\operatorname{sim}( \mu \cdot \upsilon ) = \frac{\mu ^T \cdot \upsilon}{\left \| \mu ^T \right \| \cdot \left \| \upsilon \right \| }
		\label{eq5}
	\end{equation}
	
	Similar to the query original features $f_q$, the query noise features $\widetilde f_q$ and the query fused features $\hat{f_q}$ also calculate the cluster loss with the cluster features in their respective memory banks. Therefore, the overall cluster loss is as follows:
	\begin{equation}
		L_{\text {cluster-all}} = L_{\text {cluster}} + \widetilde L_{\text {cluster}} + \hat L_{\text {cluster}}
		\label{eq6}
	\end{equation}
	
	\textbf{Consistency Contrastive Learning Loss.} To effectively train the models, we ensure the consistency between noised features and original features. Based on the feature outputs of the two network branches and their corresponding features in the memory bank, we pay attention to the consistency of query features and noise features at both the instance level and the cluster level, aiming to map them to a similar feature space as much as possible. The consistency contrastive loss is as follows:
	
	\begin{equation}
		L_{\text{consistency}} = - \operatorname{sim}\left({f_q} \cdot \widetilde m_{+} \right) - \operatorname{sim}\left({f_q} \cdot \widetilde f_q \right)
		\label{eq7}
	\end{equation}

	\subsection{Training and Inference Procedure}
	
	Taking the original features $f$ as an example, the noised features $\widetilde f$ and the fused features $\hat f$ follow the same pattern. At the initial establishment of the memory bank, we use the average of all features corresponding to each identical pseudo-label to initialize the features of a class in the memory bank, as shown in Equation~\ref{eq8}.
	\begin{equation}
		m_i = \frac{1}{num( {\operatorname{M}} (i))} \sum_{f_i \in \operatorname{M}(i)} f_i
		\label{eq8}
	\end{equation}
	where $m_i$ is the $i$-th cluster feature in the memory bank, ${\operatorname{M}}(i)$ denotes the $i$-th cluster set, and $num( \cdot )$ indicates the number of instances of a cluster. To ensure the consistency of features in the memory bank, we refer to the method used in MoCo~\cite{he2020momentum} and adopt a momentum-based approach for updating cluster features, as illustrated in Equation~\ref{eq9}.
	\begin{equation}
		m_i \longleftarrow \alpha \cdot m_i + (1 - \alpha)\cdot f_q
		\label{eq9}
	\end{equation}
	The proposed network is optimized by a joint loss function, which is formulated as:
	\begin{equation}
		L_{\text {total}} = L_{\text {cluster-all}} + L_{\text{consistency}}
		\label{eq10}
	\end{equation}
	
	\section{Experiment}
        \label{experiment}
	
	\subsection{Data and Evaluation Protocols}
	To evaluate the effectiveness of our method, we collected a set of red panda images. By scanning the microchips implanted in the red pandas, we accurately obtained their identity information. Finally, we collected a total of 3,487 images of 43 red pandas. These images were categorized into indoor and outdoor environments, with 10 red pandas having images in both environments, while the remaining 33 only had images taken indoors.
	
	For the training dataset, we selected the 10 red pandas with outdoor images, along with another 10 that only had indoor images. The images of the remaining red pandas constituted the test dataset, with specific numbers detailed in Table~\ref{tab1}. The evaluation metrics used in our experiments are the mean Average Precision (mAP) and identification accuracy at Rank-1, Rank-5, and Rank-10.
	
	\begin{table}[h]
		\caption{Red Panda Data}
		\begin{center}
			\begin{tabular}{cccccc}
				\hline
				\multicolumn{2}{c}{Train} & \multicolumn{2}{c}{Query} & \multicolumn{2}{c}{Gallery} \\
				\hline
				ID & Images & ID & Images & ID & Images \\
				\hline
				20 & 1967 & 23 & 138 & 23 & 1382 \\
				\hline
			\end{tabular}
		\end{center}
		\label{tab1}
	\end{table}
 
	\subsection{Implementation Details}
	In our method, we initialize both network branches with a ResNet-50~\cite{he2016deep} backbone encoder pre-trained on ImageNet~\cite{krizhevsky2012imagenet} as the feature extractor. After the layer-4, we remove all sub-module layers and add global average pooling, followed by a batch normalization layer~\cite{ioffe2015batch} and an L2-normalization layer, which produces features with a dimensionality of $D = 2048$. The fusion layer is a fully connected layer from $2D$ to $D$. 
 
 At the start of each epoch, we employ DBSCAN~\cite{ester1996density} for clustering to generate pseudo-labels. Input images are resized to 128 x 256~$(H \times W$). For training images, we perform random horizontal flipping, 10-pixel padding, and random cropping. The network is optimized by the Adam optimizer~\cite{kingma2014adam} with a weight decay of 0.0005 and train for 60 epochs. The initial learning rate is set to 0.00035 and is reduced by a factor of 0.1 every 20 epochs. The batch size is set to 64. The loss temperature $\tau$ in Equation~\ref{eq4} is set to 0.05, and the momentum update factor $\alpha$ is set to 0.1.
	
	\subsection{Ablation Study}
 
	\textbf{Effectiveness of feature-aware noise addition module.} In this experiment, we analyzed the effectiveness of our proposed Feature-Aware Noise Addition module. We performed feature-aware processing based on the feature maps from both convolutional and batch normalization layers. 

 \begin{figure}[h]
		\centerline{\includegraphics[width=1\linewidth]{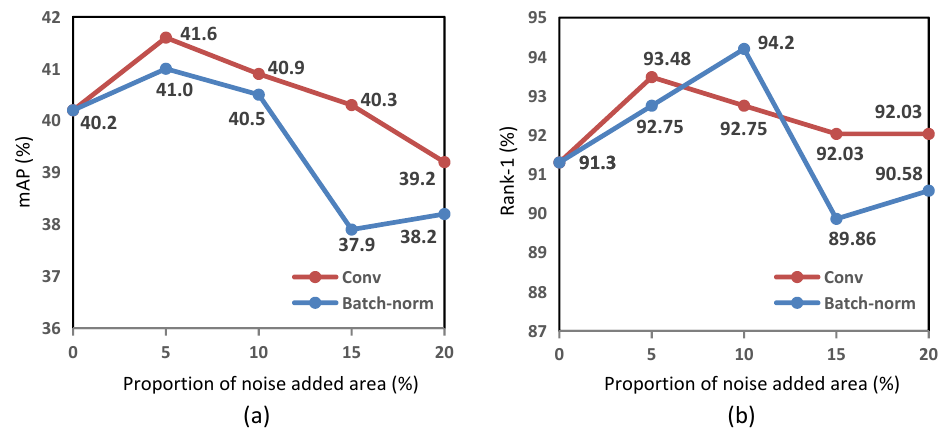}}
		\caption{(a) and (b) respectively demonstrate the changes in the model's mAP and Rank-1 metrics as the proportion of added noisy space increases. 
		}
		\label{fig3}
	\end{figure}

 	\begin{figure*}[th]
		\centerline{\includegraphics[width=1\linewidth]{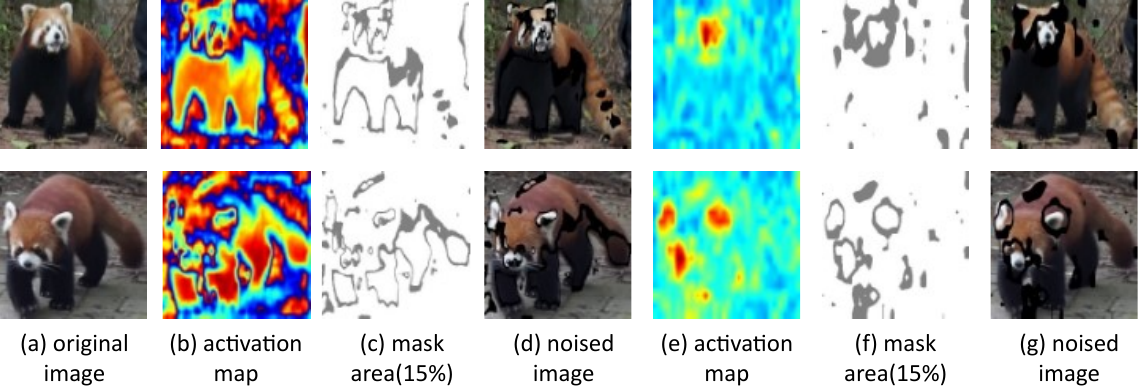}}
		\caption{Visualization of the Feature-Aware Noise Addition (FANA) module function. (a) Original image, (b) and (e) Activation maps of (a) in the convolution layer and batch normalization layer. (c) and (f) Feature-aware regions selected based on the activation maps. (d) and (g) Noised images obtained by adding noise to the corresponding regions in (a). 
		}
		\label{fig4}
	\end{figure*}
    As shown in Figure~\ref{fig3}, compared to scenarios without noise addition (0\% noise addition area), the model's accuracy improved when the noise percentage was 5\% and 10\%, with the best mAP increased by 1.4\% and the best Rank-1 increased by 2.9\%. Hence, the design of FANA helps the model to better learn the distinguishing features of each red panda. 
	
	However, we also noticed a decrease in accuracy when a larger area of features was noised. We speculate that this decrease in accuracy is likely due to the loss of a substantial amount of important features, resulting in impaired feature consistency. Therefore, it is necessary to design a smaller noise proportion like 5\% and 10\%. 
	
    \textbf{Effectiveness of Consistency loss.} In this experiment, we analyzed the effectiveness of our proposed noised feature and the original feature consistency loss. The experimental results are shown in Table~\ref{tab2}. The consistency loss is divided into two parts: one at the instance level and the other at the cluster level. 
    
    Compared with the baseline, the cluster-level loss contributes improvements of 0.73\% and 0.7\% in rank-1 and mAP, respectively. The cluster loss further enhances the performance in rank-1 and mAP by 4.35\% and 0.86\%, respectively. When using both losses, the model achieves the best performance. This indicates that the loss at both levels is meaningful. Maintaining consistency between the original features and the noised features, as well as with the cluster features in the noise memory bank, can effectively improve the performance of the encoder. 
 
    \begin{table}[h]
		\caption{Effectiveness of Consistency loss (\%)}
		\begin{center}
		\begin{tabular}{lllll}
		  \hline
		  Loss function& mAP& Rank-1 & Rank-5 & Rank-10\\
    	\hline
		  None& 38.6& 89.13& 95.65
		  &97.10\\
		  + Cluster level& 39.3& 89.86& 96.38& 97.83\\
		  + Instance level& 39.4& 93.48& 96.38&96.38\\
		  + Cluster level + Instance level& \textbf{41.6}& \textbf{93.48}& \textbf{96.38}& \textbf{97.10}
		  \\
		\hline
		\end{tabular}
		\end{center}
		\label{tab2}
	\end{table}

  \begin{table}[h]
		\caption{Comparison of Methods (\%)}
		\begin{center}
			\label{tab:method_comparison}
			\begin{tabular}{lllll}
				\hline
				\textbf{Method} & \textbf{mAP} & \textbf{Rank-1}& \textbf{Rank-5}&\textbf{Rank-10}\\
				\hline
				\multicolumn{5}{c}{Supervised Method}\\
				\hline
				DenseNet-121~\cite{huang2017densely}& 31.8
				& 81.16 & 95.65 &97.10\\
				ResNet-50~\cite{he2016deep} & 33.1& 78.26& 95.65&97.82\\
				PCB~\cite{sun2018beyond}& 40.6 & 93.78 & 97.82 &98.55\\
				\hline
				\multicolumn{5}{c}{Unsupervised Method}\\
				\hline
				SpCL~\cite{ge2020self} & 27.0
				& 78.99& 93.48&97.10
				\\
				ICE~\cite{chen2021ice} & 34.3& 83.33& 93.48
				&95.65\\
				CC~\cite{dai2022cluster} & 36.1
				& 86.96
				& 94.20
				& 94.93
				\\
				PPLR~\cite{cho2022part} & 37.6& 90.58
				& 94.93&97.10\\
				\hline
				\textbf{FANCL(Our)} & \textbf{41.6}& \textbf{93.48}& \textbf{96.38}& \textbf{97.10}
				\\
				\hline
			\end{tabular}
		\end{center}
		\label{tab3}
	\end{table}

\begin{figure*}[t]
		\centerline{\includegraphics[width=1\linewidth]{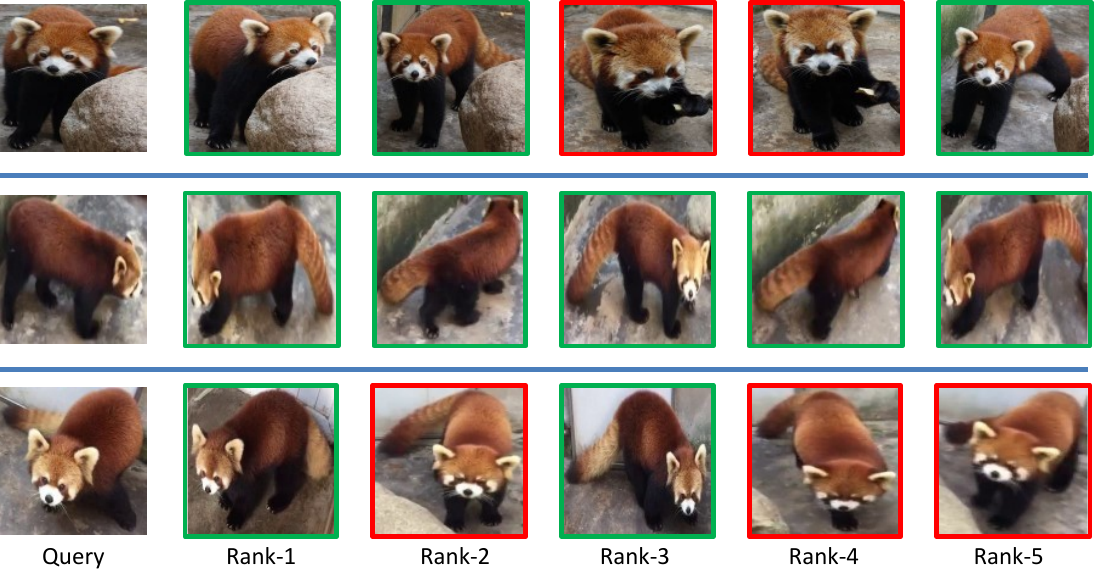}}
		\caption{Visualization of the retrieval results for example query red panda images by our proposed method. The green and red bounding boxes indicate correct and incorrect matches, respectively.
		}
		\label{fig5}
	\end{figure*}

 \subsection{Comparison with Existing Methods}
	
	We compared our proposed method with some of most advanced unsupervised re-identification methods~\cite{ge2020self,chen2021ice,dai2022cluster,cho2022part}. Additionally, we provided some supervised baselines~\cite{huang2017densely,he2016deep,sun2018beyond}, which use image labels during training to assess the effectiveness of our method. Each query image of a red panda is then matched in the gallery based on feature similarity. The results are shown in Table~\ref{tab3}. It can be seen that our FANCL method achieves significant performance improvements over other unsupervised methods, and even exhibits good performance in comparison with supervised methods.
 
	Compared to the supervised learning baseline PCB~\cite{sun2018beyond}, our method has a very slight gap in rank metrics but achieves a 1\% improvement in mAP. It is believed that this may be because the features learned by FANCL are more concentrated. This demonstrates that unsupervised methods, through careful design, can bridge the gap with supervised methods in red panda re-identification, consistent with conclusions drawn in people re-identification. Compared to other unsupervised methods, our approach also performs excellently, achieving state-of-the-art results in both mAP and rank-1 metrics. This proves that our proposed method can better guide the model in representing features related to the complex body deformations or pose variations of red pandas.

	\subsection{Visualization and Analysis}
 
	The activation maps visualized in Figure~\ref{fig4}, are generated based on the convolutional and batch norm layers in ResNet50~\cite{he2016deep}. We observe that the encoder automatically focuses on the key features of individual red pandas, with the convolutional layers almost outlining the entire silhouette of red pandas. In the process of generating noised images by adding noise to areas with high feature perception, we employ a simple random erasing method to simulate pepper noise. By adding pepper noise to key feature areas of the red panda (such as the face and tail), the resulting images not only present a more challenging task, encouraging the model to learn deeper and more robust features of the red pandas but also guide the model to learn more global and stable features through their similarity to the original images. 
 
	In Figure~\ref{fig5}, we showcase a retrieval result on the test set using the FANCL method. We notice that the model tends to find an image with the same ID among the top similar ones, yet it lacks strong concentration ability. This is reflected in the high-rank metrics and low mAP scores on the red panda dataset. We believe this is caused by the complex postures of red pandas. This not only highlights the challenges in individual recognition of red pandas but also indicates that the approach of extracting more robust and consistent features of red pandas is valid. 

    \section{Conclusions}
    \label{conclusions}
    In this paper, we explore the application of unsupervised learning methods to animal re-identification. Taking red pandas as an example, we introduce the FANCL method, where noised images with some discriminative regions contaminated are generated through a feature-aware noise addition module. By imposing the contrastive loss between cluster features and query instance features, as well as the consistency loss between original and noised features, FANCL further promotes the neural networks to learn more robust and deeper features of red pandas. We collected a set of red panda image datasets, and the experimental results prove the superiority of our proposed method while demonstrating the positive effects of the feature-aware noise addition module and consistency loss on the framework.
	
    We notice that the region selection method based on activation maps in this paper causes the problem of over-emphasis on local details of the image and ignoring the overall information, which will be the focus of our subsequent improvements. Secondly, exploring more reliable noise generation methods for animal re-ID tasks is also a meaningful research direction.

    \section*{Acknowledgements}
    This work is supported in part by the National Natural Science Foundation of China (No. 62176170), Chengdu Science and Technology Program (2022-YF09-00019-SN), the Science and Technology Department of Tibet (Grant No. XZ202102YD0018C), and the top-notched student program of Sichuan University.
    \bibliographystyle{IEEEtran}
    \bibliography{conference_101719}

\end{document}